\documentclass[conference]{IEEEtran}
\IEEEoverridecommandlockouts
\usepackage{cite}
\usepackage{amsmath,amssymb,amsfonts}
\usepackage{algorithmic}
\usepackage{graphicx}
\usepackage{textcomp}
\usepackage{xcolor}
\usepackage{amssymb}
\usepackage{mathtools}
\usepackage{booktabs}
\usepackage{balance,color}
\usepackage{multicol}
\usepackage[bookmarks=false]{hyperref}

\def\BibTeX{{\rm B\kern-.05em{\sc i\kern-.025em b}\kern-.08em
    T\kern-.1667em\lower.7ex\hbox{E}\kern-.125emX}}
\begin{document}

\title{Restricted Recurrent Neural Networks\\
\thanks{This work was supported in part by Office of Naval Research Grant No. N00014-18-1-2244. We provide our implementation at \url{https://github.com/dem123456789/Restricted-Recurrent-Neural-Networks}.}
}

\author{\IEEEauthorblockN{Enmao Diao}
\IEEEauthorblockA{\textit{Electrical and Computer Engineering}\\
\textit{Duke University}\\
Durham, USA \\
enmao.diao@duke.edu}
\and
\IEEEauthorblockN{Jie Ding}
\IEEEauthorblockA{\textit{Statistics} \\
\textit{University of Minnesota}\\
Minneapolis, USA \\
dingj@umn.edu}
\and
\IEEEauthorblockN{Vahid Tarokh}
\IEEEauthorblockA{\textit{Electrical and Computer Engineering} \\
\textit{Duke University}\\
Durham, USA \\
vahid.tarokh@duke.edu}
}

\maketitle
\IEEEpubid{978-1-7281-0858-2/19/\$31.00 \copyright\ 2019 IEEE}

\begin{abstract}
Recurrent Neural Network (RNN) and its variations such as Long Short-Term Memory (LSTM) and Gated Recurrent Unit (GRU), have become standard building blocks for learning online data of sequential nature in many research areas, including natural language processing and speech data analysis. In this paper, we present a new methodology to significantly reduce the number of parameters in RNNs while maintaining performance that is comparable or even better than classical RNNs. The new proposal, referred to as Restricted Recurrent Neural Network (RRNN), restricts the weight matrices corresponding to the input data and hidden states at each time step to share a large proportion of parameters. The new architecture can be regarded as a compression of its classical counterpart, but it does not require pre-training or sophisticated parameter fine-tuning, both of which are major issues in most existing compression techniques. Experiments on natural language modeling show that compared with its classical counterpart, the restricted recurrent architecture generally produces comparable results at about 50\% compression rate. In particular, the Restricted LSTM can outperform classical RNN with even less number of parameters.

\end{abstract}


\section{Introduction}
With the increasing volume of online streaming data generated from mobile devices, deploying efficient neural network models on smart devices with low computational power has become an emerging challenge. Recurrent Neural Network (RNN) is widely used as for online learning tasks such as time series prediction, language modeling, text generation, machine translation, speech recognition, and text to speech generation, etc. 
It has been known that Vanilla RNN fails to capture long term dependencies among long sequence of data observations.
Long-Short Term Memory (LSTM) is later developed by \cite{hochreiter1997long} in order to solve this problem by introducing memory state and multiple gating mechanism.
But this creates another practical issue since large number of states and gates require higher computational complexity and memory space. To alleviate this problem, Gated Recurrent Units (GRU) have been developed to inherit the merit of capturing long term dependencies while reducing the computational cost \cite{cho2014learning}. 
Due to the hidden state and gating mechanism, RNN, GRU and LSTM have 2x, 6x, and 8x number of parameters compared with fully connected neural networks.
Despite the huge success of applying RNNs to real-world challenges, there still exists a huge demand in reducing the model complexity to save training time, or to facilitate more economical hardware implementations. This has motivated the recent literature on deep learning model compression \cite{han2015deep,wen2016learning,cheng2017survey}. 

\IEEEpubidadjcol

In order to address these issues, we propose a new deep learning architecture called Restricted Recurrent Neural Networks (RRNN) to reduce the number of parameters by exploiting the recurrent structures. The main idea of the proposed architecture is to utilize the sharing of model parameters corresponding to the input and hidden states while maintaining comparable or even better performance. Classical RNNs assign completely separate weight matrices for the input and hidden states at each gate. The underlying assumption is that those weight matrices can be trained to adapt to their corresponding input. An alternative case is where all input and hidden states at each gate share exactly the same weight matrix. This may lead to undesirable results because a single weight matrix cannot be easily trained to accommodate two distinct distributions of inputs. Nevertheless, it is reasonable to assume certain dependencies between the input and hidden states in the context of sequential models with recurrent structure. Our intuition is that we can utilize shared parameters to account for the dependencies between two inputs while granting a portion of specialized free parameters for each distinct source of inputs.  Interestingly, our experiments show that  the idea of enforcing parameter sharing  works quite well for RNNs. 
By sharing most of the model parameters, we can reduce the model complexity while still producing comparable or even better performance than classical models. 

The proposed solution can be regarded as a new strategy of model compression. 
Classical model compression techniques mainly fall into two categories, namely parameter pruning or quantization, and low-rank factorization \cite{cheng2017survey}. Parameter pruning or quantization typically explores the redundancy of model parameters and removes non-informative neurons after pre-training the model. On the other hand, matrix low-rank factorization recognizes RNN as a combination of multiple 2D matrices and decompose them \cite{prabhavalkar2016compression,lu2016learning,tjandra2017compressing,grachev2019compression}. However, the practical use of both methods are limited by computational disadvantages. Pruning with $\mathcal{L}_1$ or $\mathcal{L}_2$ regularization requires more iterations to converge and sophisticated fine-tuning of the regularization parameters. Low-rank factorization typically requires computationally expensive decomposition operations, and to achieve desirable performance comparable with classical RNNs, both methods typically require retraining from a pre-trained model. 
In fact, the parsimonious architecture can be regarded as a compression of its classical counterpart, but with the following advantages. 

The main contribution of this paper is as follows. First, we propose a novel model compression technique specifically taking advantage of the recurrent structures of RNNs. Unlike classical model compression techniques, our method does not require retraining a pre-trained model. 
Second, our method can explicitly control the compression rates, whereas deep compression based on regularization typically cannot directly relate regularization parameters to the exact compression rates. 
Third, since we only alter the structure of the weight matrix, our method is compatible with the existing regularization techniques such as Dropout \cite{srivastava2014dropout}. Finally, our numerical experiments also show that sharing model parameters also regularizes the model and produce better performance when the compression rate is small.

The paper is outlined as below. We review the background and introduce the notation in Section~\ref{sec_related}. Our proposed method is introduced in Section~\ref{sec_propose}. Experimental results are given in Section~\ref{sec_experiment}. Finally, we make our conclusions in Section~\ref{sec_conclusion}.

\section{Related Work} \label{sec_related}

In spite of extensive research on compressing general deep learning models \cite{cheng2017survey}, 
little attention has been paid to particular compression techniques for Recurrent Neural Networks. 
Existing works on compressing RNN mostly focus on the application of matrix decomposition to the internal
weight matrix~\cite{tjandra2017compressing,grachev2019compression}. To some extent they ignore the underlying recurrent structure and sequential nature of the data.
Recently, there has been a variety of Convolutional Neural Networks (CNN) which enables efficient compression from the perspective of network structural design rather then sophisticated post-processing after pre-training (that is typically employed in classical compression techniques~\cite{howard2017mobilenets,zhang2018shufflenet,huang2018condensenet}).
The success of parsimonious CNN structures has motivated our work to introduce a new framework specifically designed for RNN. We depart from the existing model compression techniques in RNN, and propose a structure based approach, in the sense that our proposed method does not require the pre-training stage, and data-driven pruning of weights or neurons as the second stage. This enables simpler training, and easier hardware implementation.

\subsection{Recurrent Neural Networks}
RNN has been widely used in modeling online streaming data such as natural language texts and speech \cite{sak2014long,zaremba2014recurrent}. 
Its vanilla version can be simply written as
\begin{align}
h_t &= \tanh(W_{xh}x_t + b_{ih} + W_{hh}h_{t-1} + b_{hh})
\end{align}
where the subscript $t, x, h$ denote the time step, input layer, and hidden layer respectively.
Vanilla RNN is known to suffer from gradient vanishing and explosion problems, and it fails to capture long term dependencies among sequential data.
LSTM is later developed as a solution to the drawbacks of vanilla RNN \cite{hochreiter1997long}. As shown below, LSTM has an internal memory cell $c_t$ to store the long term dependencies. It also introduces four gates in order to to control the information flow from input, hidden state and the memory cell. The input gate $i_t$ determines how much information from input and hidden state should be remembered by the memory cell. A forget gate $f_t$ determines how much long term memory should be saved in the next time step. $g_t$ contains the information of current input and previous hidden state. The output gate $o_t$ controls the information flows into the next hidden state. Depending on the switching of gates, LSTM can represent long-term and short-term dependencies of sequential data simultaneously.
\begin{align*}
i_t &= \sigma(W_{xi}x_t + b_{ih} + W_{hi}h_{t-1} + b_{hi}) \\
f_t &= \sigma(W_{xf}x_t + b_{if} + W_{hf}h_{t-1} + b_{hf}) \\
g_t &= \tanh(W_{xg}x_t + b_{ig} + W_{hg}h_{t-1} + b_{hg})  \\ 
o_t &= \sigma(W_{xo}x_t + b_{io} + W_{ho}h_{t-1} + b_{ho}) \\
c_t &= f_t*c_{t-1}+i_t*g_t \\
h_t &= o_t*\tanh(c_t)
\end{align*}
Despite its merit of capturing long term dependencies, LSTM has four times the number of parameters of vanilla RNN and an extra memory cell. This leads to heavy computational costs and frequent overfitting issues due to its superfluous degrees of freedom.\\
These potential drawbacks popularize GRU, a lighter version of LSTM. The formulation of GRU is shown below \cite{cho2014learning}. The reset gate $r_t$ decides whether to ignore the information propagated from previous hidden states. The update gate $z_t$ couples the input and the forget gate in LSTM into one single gate to control the information flowing to the next hidden state. Compared with LSTM, GRU is computationally more efficient because no memory cell is used and its coupling techniques reduce two weight matrices.
\begin{align*}
r_t &= \sigma(W_{xr}x_t + b_{ir} + W_{hr}h_{t-1} + b_{hr}) \\
z_t &= \sigma(W_{xz}x_t + b_{iz} + W_{hz}h_{t-1} + b_{hz}) \\
n_t &= \tanh(W_{xn}x_t + b_{in} + r_t*(W_{hn}h_{t-1} + b_{hn})) \\
h_t &= (1-z_t)*n_t + z_t*h_{t-1}
\end{align*}
Apart from LSTM and GRU mentioned above, there still exist many other variations of RNN. Many of them are designed to specialize on specific type of sequential data \cite{jozefowicz2015empirical,tai2015improved,greff2016lstm,kent2019performance}. Although our numerical experiments are based on RNN, LSTM and GRU, it is straightforward to utilize our method in other recurrent models.

\subsection{Model Compression}
As neural networks with higher layers and parameters have been achieving state-of-the-art performance in many real world challenges, reducing their computational and storage cost becomes critical. Recent advances in internet of things and wireless sensor networks have been driving an ever-increasing volume of online streaming data. \textit{Big data on small devices} has motivated deploying deep learning systems on devices with very limited computational resources. Classical model compression techniques like parameter pruning and low-rank factorization have been introduced to deep learning models \cite{cheng2017survey}. Although these methods have shown to be feasible, they usually require model pre-training and sophisticated retraining in order to achieve desirable results. 
Moreover, these techniques demand cumbersome fine-tuning of hyper-parameters. 
Another active direction of model compression is to design new architectures which are inherently more efficient than the standard models. For example, group convolutions are adopted by \cite{howard2017mobilenets,zhang2018shufflenet,huang2018condensenet} to reduce the number of parameters of vanilla CNN. Motivated by this thread, we propose a novel compression methodology specifically for RNN by taking advantage of its recurrent structure.

\section{Method} \label{sec_propose}
In this section, we first describe the restriction method that reduces the number of parameters of neural networks by enforcing parameter sharing. We will elaborate the formulations of Restricted RNN, LSTM and GRU. The degrees of freedom of Restricted RNN can be easily calculated based on our formulations.

\subsection{Restricted Recurrent Neural Networks (RRNN)}
At each time step $t$, $x_t$ and $h_{t-1}$ serve as two inputs to RNN. Suppose that the size of $x_t$ and $h_{t-1}$ are $k_{xh} \times n$ and $k_{hh} \times n$ respectively. Classical methods employ separate parameter matrices, say $W_{xh}$ and $W_{hh}$, for each type of input. An alternative way of modeling is to restrict the two inputs to use exactly the same parameter matrix, i.e. $W_{xh} = W_{hh}$ assuming the hidden channel size is the same as the input channel size. 
Intuitively speaking, the hidden states from the last time step contribute equally with the instantaneous data input. This is because $h_t$ is a function of 
$W_{xh}(x_t + h_{t-1})$. 
While the equal weighting may not be too extreme to be practically useful, it may help to consider the scenario that lies between the above two extreme cases, namely to let $W_{xh}$ and $W_{hh}$ share part of the parameters. Intuitively, this reduces overfitting by enforcing parameter sharing while allocating sufficient degrees of freedom to either type of input. By assuming that the inputs are not totally independent with each other, one could imagine shared parameters are able to capture the similarities among the inputs while non-shared parameters grant enough degrees of freedom for innovations.

\begin{figure*}[htbp]
\begin{center}
\includegraphics[width=0.8\linewidth]{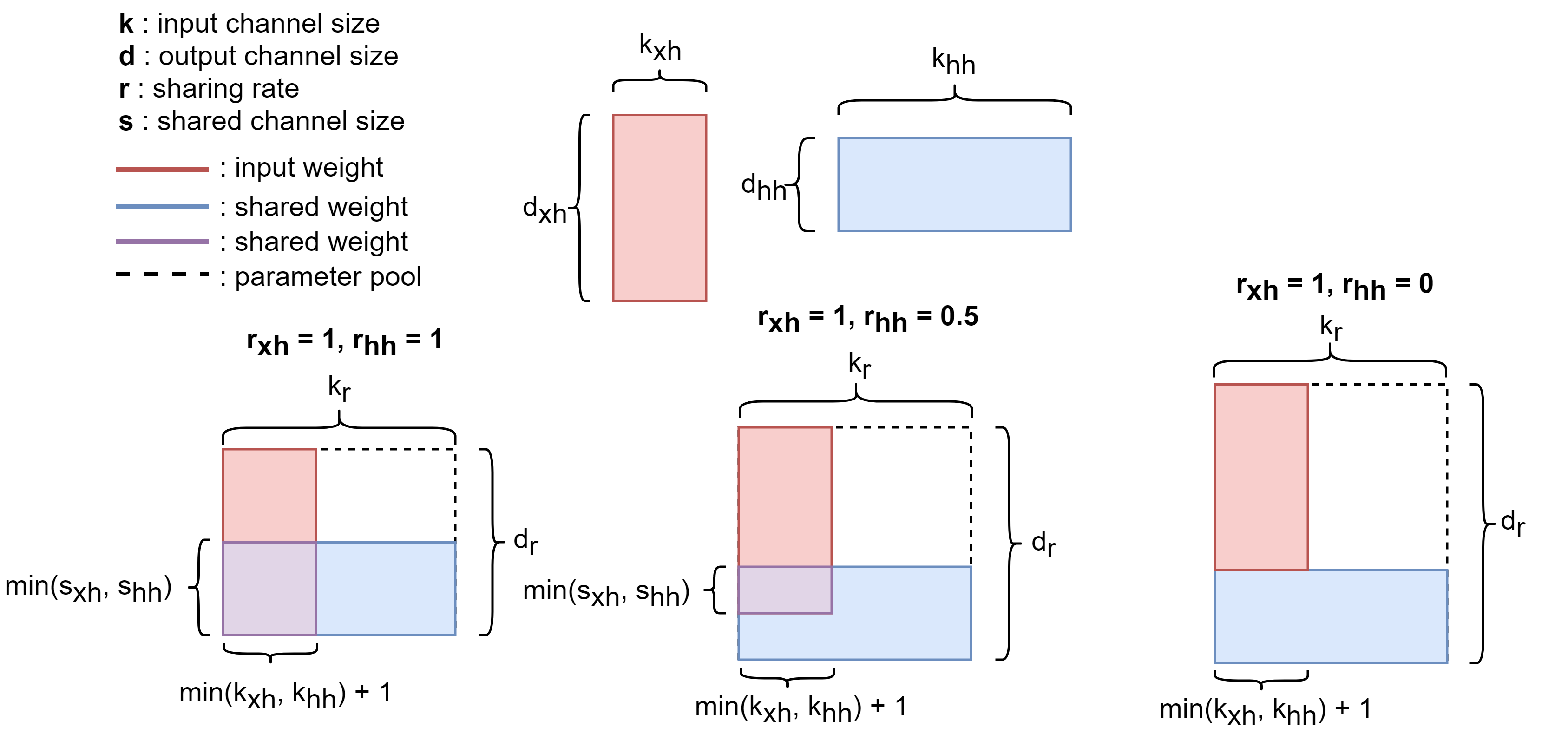}
\end{center}
	\vspace{-0.0cm}
   \caption{Illustration of parameter restriction in RRNN.}
   \vspace{-0.0cm}
\label{fig_illustrate}
\end{figure*}

Specifically, suppose that the size of $W_{xh}$ and $W_{hh}$ are $d_{xh} \times k_{xh}$ and $d_{hh} \times k_{hh}$ respectively. Since we need to sum up $W_{xh}x_t$ and $W_{hh}h_{t-1}$, the common practice is that $d_{xh}$ is equal to $d_{hh}$. Suppose that $r_{xh} \in [0,1]$ and $r_{hh} \in [0,1]$ are sharing rates. Then the restricted parameter matrices $W^{r}_{xh}$ and $W^{r}_{hh}$ are formulated as below. 
Additionally, for illustration, a diagram is provided in Figure~\ref{fig_illustrate}.
\begin{align*}
s_{xh} &= Round(r_{xh}\times d_{xh}),\, s_{hh} = Round(r_{hh}\times d_{hh})\\
q_{xh} &= d_{xh}-s_{xh},\, q_{hh} = d_{hh}-s_{hh},\\ 
s_{r} &= max(s_{xh},s_{hh}),\, k_{r} = max(k_{xh},k_{hh})\\
d_{r} &= s_{r} + q_{xh} + q_{hh}	
\end{align*}
\begin{align*}
W &\sim (d_{r},\, k_{r}),\, b \sim (d_{r},)\\
W^{r}_{xh} &=\begin{pmatrix}
    W[:s_{xh},:k_{xh}]\\
    W[s_{r}:s_{r}+q_{xh},:k_{xh}]
    \end{pmatrix}\\
b^{r}_{xh} &= \begin{pmatrix}
    b[:s_{xh}]\\
    b[s_{r}:s_{r}+q_{xh}]
    \end{pmatrix}\\
W^{r}_{hh} &= \begin{pmatrix}
    W[:s_{hh},:k_{hh}]\\
    W[s_{r}+q_{xh}:s_{r}+q_{xh}+q_{hh},:k_{hh}]
    \end{pmatrix}\\
b^{r}_{hh} &= \begin{pmatrix}
    b[:s_{hh}]\\
    b[s_{r}+q_{xh}:s_{r}+q_{xh}+q_{hh}]
    \end{pmatrix}\\
h_t &= \tanh(W^{r}_{xh}x_t + b^{r}_{xh} + W^{r}_{hh}h_{t-1} + b^{r}_{hh})
\end{align*}
$s_{ih}$ and $s_{hh}$ represent the output channel size of shared parameters for the input and hidden state. We first allocate parameter pools $W$ and $b$ for the restricted parameters to index from. The parameter $k_{r}$, the input channel size of $W$, is equal to the maximum of input channel size of input and hidden state. The parameter $d_{r}$, the output channel size of $W$, is equal to the summation of the maximum shared output channel size and non-shared channel size $q$ for each state. Restricted parameters for each state, $W^{r}_{ih}$ and $W^{r}_{hh}$, are selected from parameter pools $W$ with indices that enforce parameter sharing among the input and hidden state. Restricted parameters are concatenated from the shared and non-shared sub-matrices indexed from parameter pools (i.e. $W$ and $b$). Those indices also make sure the size of our restricted model exactly match its non-restricted counterpart (which does not employ sharing). Therefore, it is straightforward to replace non-restricted parameters in RNN with their restricted versions. It is worth mentioning that non-indexed parameters in $W$ are not involved in training and only treated as placeholders for notational convenience. The trainable parameters involved in computation are $W^{r}_{\cdot}$ and $b^{r}_{\cdot}$. The number of shared parameters $S_{r}$ and the number of overall parameters $P_{r}$ of Restricted RNN can be computed as follows.
Let
\begin{align*}
d &\coloneqq d_{xh} = d_{hh}\\
P_{xh} &= d \times (k_{xh}+1)\\
P_{hh} &= d \times (k_{hh}+1)\\
P &= P_{ih} + P_{hh}\\
S_{r} &= min(s_{xh},s_{hh}) \times (min(k_{xh},k_{hh})+1)\\
P_{r}& = P - S_{r}
\end{align*}
Then compression rate $C$ can be calculated by 
$$C = \frac{P_{r}}{P}$$
Assuming the common practice that $k \coloneqq k_{xh}=k_{hh}$, $r\coloneqq r_{xh}=r_{hh}$, 
we have 
\begin{align*}
s &\coloneqq s_{xh} = s_{hh} = r \times d\\
P &= 2d \cdot (k+1),\, S_{r} = s \cdot (k+1)
\end{align*}
The compression rate $C$ can be simplified as follows.
\begin{align*}
C &= \frac{P_{r}}{P} = \frac{P-S_{r}}{P} = \frac{2d-s}{2d} = \frac{2-r}{2}
\end{align*}
It can be seen that $C \in [0.5,1]$. In fact, $C=0.5$ corresponds to the extreme case that input and hidden state share all the parameters, while $C=1$ corresponds to the classical case without any sharing.

\subsection{Generalized RRNN for LSTM and GRU}
Since our method is designed to take advantage of the recurrent structures, it can also be applied to other variations of RNN such as LSTM and GRU. The distinction between these more complex recurrent models and vanilla RNN is the gating mechanism. As shown in Section 2, the gating mechanism simply replaces tanh with sigmoid activations functions as a soft switch to control the information flow. Naturally, we can also enforce parameter sharing among gates through our restriction method. The formulations of Restricted LSTM and GRU are analogous to Restricted RNN. Here we generalize our previous formulation of Restricted RNN to handle multiple inputs and outputs. This is common in other variations of RNN like Tree-LSTM which can have more inputs than input and hidden state \cite{tai2015improved}. For example, RNN, LSTM and GRU all have two inputs including the input and hidden state. They also have one, four and three outputs representing different gates respectively. Given $m$ inputs and $n$ outputs, we can generalize our formulation as follows. The matrix $s_{m \times n}$ indicates that each pair of input and output can have its own sharing rate. The non-linearity $f_n$ can be either tanh or sigmoid function depending on the structure of RNN.
\begin{align*}
s_{m \times n} &= Round(r_{m \times n}\times d_{m \times n})\\
q_{m \times n} &= d_{m \times n}-s_{m \times n}\\ 
s_{r} &= max(s_{m \times n}),\, k_{r} = max(k_{1:m})\\
d_{r} &= s_{r} + \sum_{i=1}^{m}\sum_{j=1}^{n}q_{ij}	
\end{align*}
\begin{align*}
W &\sim (d_{r},\, k_{r}),\, b \sim (d_{r},)\\
W^{r}_{mn} &=\begin{pmatrix}
    W[:s_{mn},:k_{m}]\\
    W[s_{r}+\sum_{\substack{1\leq i \leq m \\ 1 \leq j \leq n-1}} q_{ij}:s_{r}+\sum_{\substack{1\leq i \leq m \\ 1 \leq j \leq n}} q_{ij},:k_{m}]
    \end{pmatrix}\\
b^{r}_{mn} &= \begin{pmatrix}
    b[:s_{mn}]\\
    b[s_{r}+\sum_{\substack{1\leq i \leq m \\ 1 \leq j \leq n-1}} q_{ij}:s_{r}+\sum_{\substack{1\leq i \leq m \\ 1 \leq j \leq n}} q_{ij}]
    \end{pmatrix}\\
y^{n}_{t} &= f_{n}(\sum_{i=1}^{m}W^{r}_{mn}x^{m}_{t}+b^{r}_{mn})
\end{align*}
The number of parameters of Generalized RRNN can be computed as follows.
\begin{align*}
P_{mn} &= d_{mn} \times (k_{m}+1)\\
P &= \sum_{i=1}^{m}\sum_{j=1}^{n}P_{ij}\\
S_{r} &= (mn-1) \times min(s_{m \times n}) \times (min(k_{1:m})+1)\\
P_{r}& = P - S_{r}
\end{align*}
By assuming all sharing rates, input and output channel size are the same size and no rounding is performed, the compression rate $C$ can be simplified as follows.
\begin{align*}
P &= mnd(k+1)\\
C &= \frac{P_{r}}{P} = \frac{P-S_{r}}{P} = \frac{mnd-(mn-1)s}{mnd} \approx 1 - \frac{s}{d} = 1 - r
\end{align*}

\begin{table*}[tb]
\centering
\caption{Comparison with state-of-the-art architectures in terms of Test Perplexity on Penn Treebank dataset}
\label{tab:table0}
\begin{tabular}{@{}ccc@{}}
\toprule
Model                  & Model parameters (M)        & Test Perplexity \\ \midrule
LR LSTM 200-200\cite{grachev2019compression}        & 0.928                       & 136.115         \\
LSTM-SparseVD-VOC\cite{chirkova2018bayesian}      & 1.672                       & 120.2           \\
KN5 + cache\cite{mikolov2012context}            & 2                           & 125.7           \\
LR LSTM 400-400\cite{grachev2019compression}        & 3.28                        & 106.623         \\
LSTM-SparseVD\cite{chirkova2018bayesian}          & 3.312                       & 109.2           \\
RNN-LDA + KN-5 + cache\cite{mikolov2012context} & 9                           & 92              \\
AWD-LSTM\cite{merity2017regularizing}               & 22                          & 55.97           \\ \midrule
\textbf{RLSTM-Tied-Dropout} (r=0.5)       & \textbf{2} (Embedding) + \textbf{0.553} (RNN) & \textbf{103.5}           \\ \bottomrule
\end{tabular}
\end{table*}

\vspace{0.4cm}
\section{Experiments} \label{sec_experiment}

\subsection{Experimental setup}
To demonstrate the effectiveness of our RRNN methodology, we conduct sequential prediction experiments on the Penn Treebank (PTB) dataset and the WikiText-2 (WT2) dataset \cite{mikolov2010recurrent,merity2016pointer}. The Penn Treebank dataset is a well-recognized benchmark dataset for language modeling experiments. The language model is supposed to make prediction of the next word based on the previous text. The dataset has 10k words in its vocabulary and consists of 929k training words, 73k validation words, and 82k test words. It does not contain capital letters, numbers, or punctuation which constitute 5.8\% out of vocabulary (OoV) tokens. WikiText-2 contains sentences sourced from Wikipedia articles and is more challenging compared with PTB dataset. The dataset contains a vocabulary of 33,278 words and includes 2M training words, 210K validation words, and 240K test words. Unlike PTB dataset, it retains capitalization, punctuation, and numbers and has 2.6\% OoV tokens.

A typical neural network architecture used in language modeling consists of an embedding layer, recurrent layers and a softmax layer. Both embedding layer and softmax layer are fully connected neural networks. 
We experiment with three recurrent layers with 200 hidden units and 200 embedding size.
To train our models, we minimize cross entropy loss averaged over all words with Stochastic Gradient Descent (SGD) optimizer of 0.9 momentum and 1e-6 weight decay. We carry out gradient clipping with maximum norm 0.25 and an initial learning rate of 1 for training 100 epochs. We use cosine annealing learning rate as it empirically produces much more stable results than using a holdout validation set \cite{loshchilov2016sgdr}. We use a batch size 80 and 35 Back Propagated Through Time (BPTT) length for both datasets. We evaluate all our models against two quantitative metrics, \textit{perplexity} and the \textit{number of free model parameters}. The perplexity, which is the exponentiation of the cross entropy, is a classical metric for language modeling. It evaluates the uncertainty of words predicted by a model. High perplexity means that the model produces near-uniform random predictions from the vocabulary, and thus is undesirable.

To demonstrate the tradeoff between perplexity versus the number of model parameters, we apply our restriction method with RNN, LSTM and GRU at different sharing rates $r=\{0,0.1,0.3,0.5,0.7,0.9,0.95,1\}$. We also demonstrate the result of models regularized by tied embedding and Dropout with 0.2 rate \cite{press2016using}.

\subsection{Results}

The main results of our compressed models are summarized in Table~\ref{tab:table0}. We compare our results to several state-of-the-art recurrent models for language modeling which either use a small number of parameters or is compressed from a larger model. We are unable to find baseline results on WT2 dataset, so we only compare over PTB dataset. 

\begin{table}[htbp]
\centering
\caption{Model Complexity of our proposed RRNN and its variants (unit: million)}
\label{tab:table1}
\begin{tabular}{@{}cccclllll@{}}
\toprule
r & 1     & 0.95  & 0.9   & 0.7   & 0.5   & 0.3   & 0.1   & 0     \\ \midrule
RRNN          & 0.130 & 0.136 & 0.142 & 0.167 & 0.191 & 0.215 & 0.239 & 0.251 \\
RGRU          & 0.130 & 0.161 & 0.191 & 0.311 & 0.432 & 0.553 & 0.673 & 0.733 \\
RLSTM         & 0.130 & 0.173 & 0.215 & 0.384 & 0.553 & 0.721 & 0.890 & 0.975 \\ \bottomrule
\end{tabular}
\end{table}

\begin{table*}[htbp]
\centering
\caption{Test (Validation) Perplexity of the proposed architectures (denoted by prefix `R') on Penn Treebank dataset}
\label{tab:table2}
\begin{tabular}{@{}ccccccccc@{}}
\toprule
r & 1             & 0.95          & 0.9           & 0.7           & 0.5           & 0.3           & 0.1           & 0             \\ \midrule
RRNN          & 188.9 (197.9) & 208.6 (219.5) & 200.4 (211.8) & 175 (183.9)   & 176.5 (186.7) & 156.9 (165.1) & 156.6 (164.4) & 154.8 (162.4) \\
RGRU          & 154.2 (162)   & 153.7 (160)   & 153.5 (160.9) & 152.4 (157.4) & 148.4 (154.9) & 148.1 (153)   & 146.6 (151.5) & 144.9 (150.3) \\
RLSTM         & 188.2 (199.5) & 159.9 (167.8) & 148.1 (155.8) & 133.6 (139.8) & 129 (133.9)   & 127.9 (133.8) & 123.8 (129.8) & 124.6 (130.4) \\
RRNN-Tied-Dropout       & 244.8 (252)   & 242.1 (248.7) & 239.8 (246.6) & 231.2 (238.7) & 226.6 (233.7) & 224.3 (229.9) & 223.3 (230)   & 221.7 (228)   \\
RGRU-Tied-Dropout       & 199 (207.8)   & 190.3 (198.5) & 184.2 (191.1) & 169.8 (175.9) & 163.6 (168.6) & 161.6 (168.3) & 162.2 (168)   & 156 (162.8)   \\
RLSTM-Tied-Dropout      & 188.5 (197)   & 141.2 (148.6) & 126.7 (131.8) & 106.6 (111.4) & \textbf{103.5 (108.6)} & 104.1 (109.7) & 105.8 (110.7) & 107.7 (112.5) \\ \bottomrule
\end{tabular}
\end{table*}

\begin{table*}[htbp]
\centering
\caption{Test (Validation) Perplexity of the proposed architectures (denoted by prefix `R') on WikiText2 dataset}
\label{tab:table3}
\begin{tabular}{@{}ccccccccc@{}}
\toprule
r & 1             & 0.95          & 0.9           & 0.7           & 0.5           & 0.3           & 0.1           & 0             \\ \midrule
RRNN          & 289.5 (314.6) & 293.5 (319.3) & 283.5 (308.8) & 257.3 (280.4) & 249.6 (272)   & 242.1 (264.1) & 240.5 (262.9) & 230.8 (253.1) \\
RGRU          & 177.8 (193.2) & 175.1 (189.3) & 177.7 (191.9) & 174.9 (189.8) & 169.6 (187.3) & 169.6 (184.6) & 170.9 (185.8) & 167.5 (181.8) \\
RLSTM         & 181.2 (196.1) & 174.5 (187.6) & 170.9 (182.9) & 162.8 (175.6) & 158.7 (171.2) & 154.3 (167.5) & 151.8 (165)   & 154.5 (167.6) \\
RRNN-Tied-Dropout       & 303.8 (331.8) & 304.2 (332.7) & 300.2 (328.1) & 288.7 (316.2) & 283.9 (311.3) & 279.5 (305.7) & 279.1 (305.4) & 276 (302.3)   \\
RGRU-Tied-Dropout       & 228.5 (249.8) & 215.7 (235)   & 205.8 (223.7) & 179.6 (195)   & 168.1 (182.4) & 161.2 (174.5) & 154.3 (167)   & 150.4 (163)   \\
RLSTM-Tied-Dropout      & 334.4 (356)   & 170.9 (184.8) & 152.3 (164.4) & 128.8 (137.1) & 119.1 (126.4) & 114.7 (121.6) & \textbf{112.6 (118.4)} & 112.9 (119.4) \\ \bottomrule
\end{tabular}
\end{table*}

\begin{figure*}[htbp]
\begin{center}
\includegraphics[width=1\linewidth]{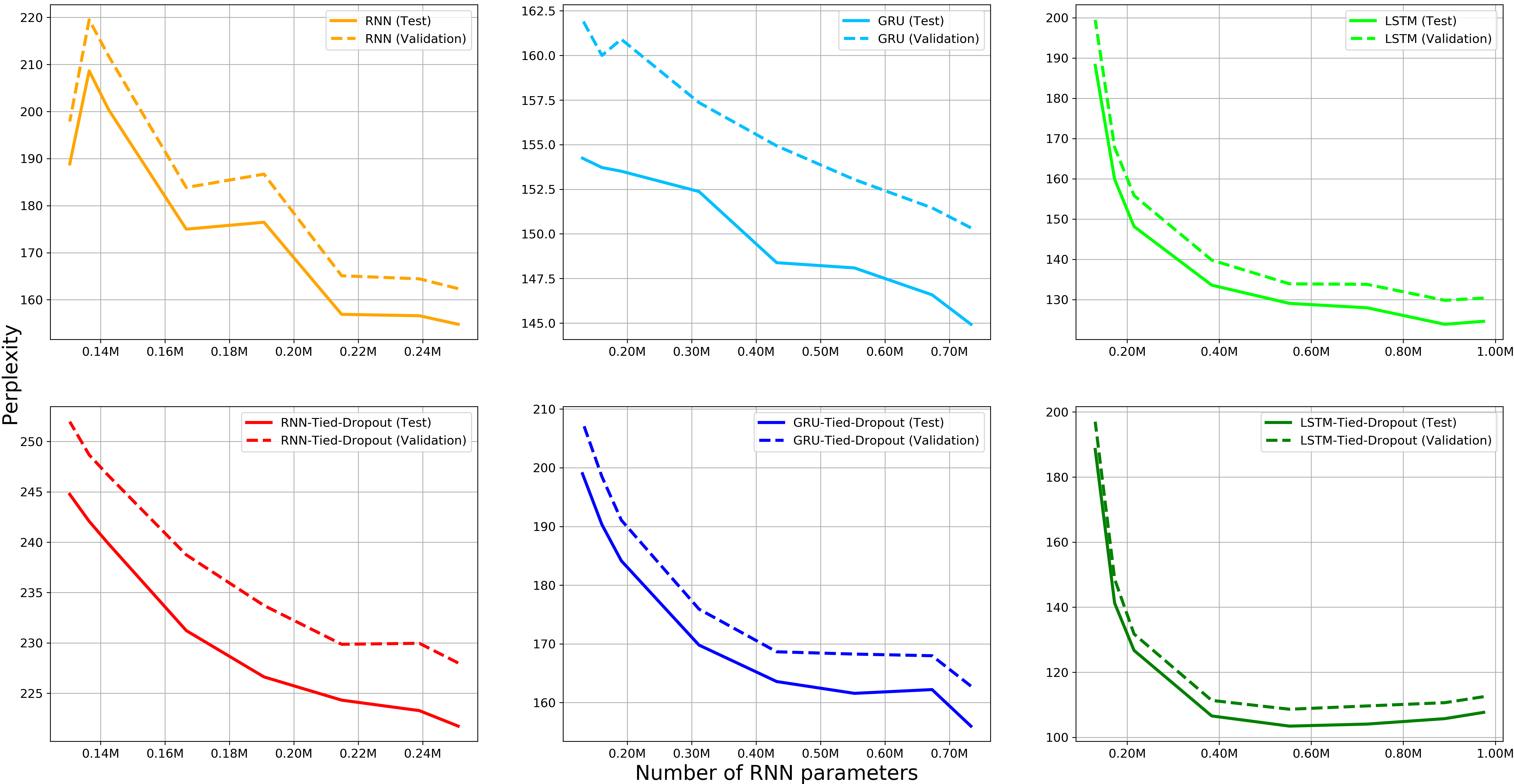}
\end{center}
	\vspace{-0.0cm}
   \caption{Perplexity vs. Number of RNN parameters for Penn Treebank dataset.}
   \vspace{-0.0cm}
\label{fig_0}
\end{figure*}

\begin{figure*}[htbp]
\begin{center}
\includegraphics[width=1\linewidth]{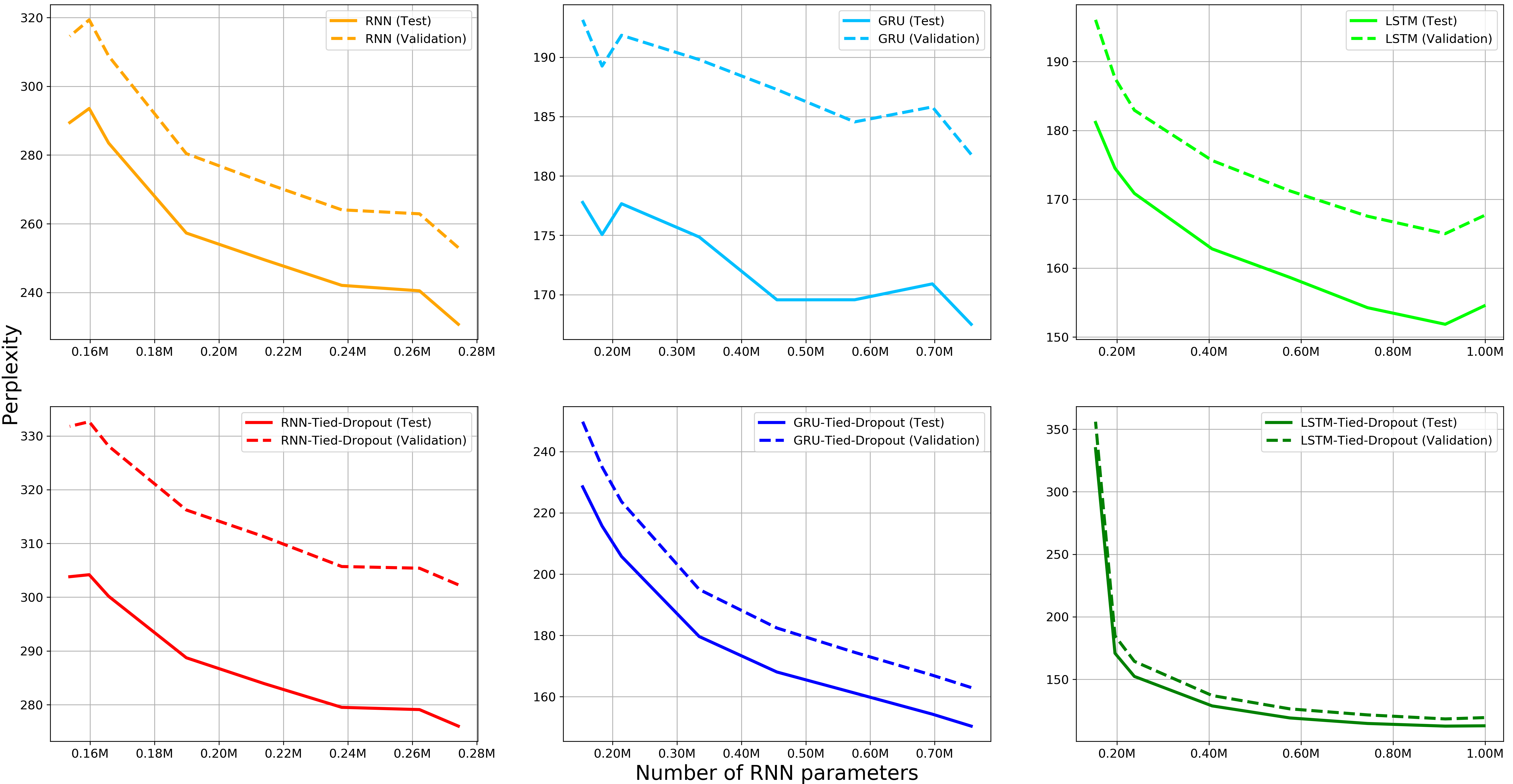}
\end{center}
	\vspace{-0.0cm}
   \caption{Perplexity vs. Number of RNN parameters for WikiText2 dataset.}
   \vspace{-0.0cm}
\label{fig_1}
\end{figure*}

We summarize the model complexities of all of our trained models in Table~\ref{tab:table1}. Since we focus on compressing RNN instead of embedding layers, we also include the number of RNN parameters in addition to overall model parameters. We demonstrate the performance of our restricted models on both PTB and WT2 datasets at different sharing rates $s$ in Table \ref{tab:table2} and \ref{tab:table3}. From the experimental results, we discovered an interesting phase transition from all-parameters-sharing to partial-parameter-sharing (particularly for Restricted LSTM). To demonstrate it, we graphically illustrate perplexity versus the number of RNN parameters in Figures~\ref{fig_0} and \ref{fig_1}.

In Table \ref{tab:table0}, we show the best known results of compressed RNN for the PTB dataset. Due to the large vocabulary size, LSTM-SparseVD and LR LSTM not only focus on compressing RNN layers but also on embedding and softmax layers. However, our restricted models only focus on compressing RNN layers. Therefore, to maintain a reasonable comparison, we train our models with tied embedding where the weight matrices of embedding and softmax layers are completely shared \cite{press2016using}. Tied embedding is a common regularization technique which prevents the model from learning a one-to-one correspondence between the input and output \cite{inan2016tying}. 
We also empirically add Dropout with rate 0.2 to regularize our models. Through various experiments we found that the final result is not very sensitive to small Dropout rates.

\balance

We illustrate the model complexity of Restricted RNN, GRU and LSTM in Table \ref{tab:table1}. As described in Section \ref{sec_propose}, sharing rate $s=1$ will force input and hidden states at all gates to share the same parameters while $s=0$ indicates classical RNN where each each input and hidden state at each gate corresponds to separate parameters. It is worth noting that it is possible to compress LSTM to a size smaller than that of vanilla RNN. In fact, our detailed results shown in Table \ref{tab:table2} and \ref{tab:table3} demonstrate that RLSTM (RLSTM-Tied-Dropout) with $s=0.9$ has lower model complexity and also outperforms vanilla RNN (RNN-Tied-Dropout). Moreover, RLSTM (RLSTM-Tied-Dropout) with $s=0.3$ significantly outperforms vanilla GRU (GRU-Tied-Dropout) when both have similar model complexity. It suggests that the merit of RNN and GRU is their faster training time, due to their less complicated recurrent structure. However, to reduce the model complexity, we should exploit the dependencies and enforce parameter sharing among input and hidden states at each gate.

As mentioned in Section \ref{sec_propose}, the intuition behind our method is that either sharing all parameters or no parameters is not the optimal modeling for multiple dependent inputs. Figure \ref{fig_0} and \ref{fig_1} show that there exists a phase transition of parameter sharing in Restricted RNN. No parameters sharing apparently produces inferior result since the model parameters are learned to make a compromise among inputs. Thus, the outputs fails to address the distinction among the inputs. Sharing all parameters in most cases produces ideal result. However, it suffers from potential overfitting and low efficiency. As shown in Figures \ref{fig_0} and \ref{fig_1}, performance improvement with sharing rate $r \geq 0.5$ tends to be much more significant than the improvement with $r<0.5$. It means that we only need to grant a small portion of degrees of freedom for each input to achieve a result which is comparable to classical settings. This phenomenon is obvious for RLSTM-Tied-Dropout. In fact, as shown in Tables \ref{tab:table2} and \ref{tab:table3}, RLSTM with small sharing rate also regularizes its classical counterparts which have more parameters and more complex recurrent structure than that of RNN and GRU.
As to the choice of $r$ in practice, better performance-compression tradeoff can be obtained by fine-tuning $r$ in specific data domains. From various experimental studies,  we suggest to set $r=0.5$ as the default option.

\section{Conclusion} \label{sec_conclusion}
In this work, we propose a novel model compression methodology called Restricted Recurrent Neural Networks (RRNN). Unlike pruning weights and decomposing estimated parameter matrices, our structure based model does not require pre-training and fine-tuning of pre-trained models. Our method explicitly takes the advantage of the recurrent structures of RNN by enforcing parameter sharing among the input and hidden state. Our work can be generalized to compress gated variations of RNN like LSTM and GRU. Our results show that both extreme cases of sharing all and none parameters are not the optimal solution to model multiple dependent input data. Sharing partial parameters can exploit the dependencies among inputs and greatly reduce the number of RNN parameters. 

\ifCLASSOPTIONcaptionsoff
  \newpage
\fi

\clearpage

\balance
\bibliographystyle{IEEEtran}
\bibliography{RRNN}

\end{document}